\documentclass[11pt]{article}
	
\newcommand{\blind}{0}

\makeatletter
\renewcommand\section{\@startsection {section}{1}{\z@}%
                                   {-3.5ex \@plus -1ex \@minus -.2ex}%
                                   {2.3ex \@plus.2ex}%
                                   {\normalfont\fontfamily{phv}\fontsize{16}{19}\bfseries}}
\renewcommand\subsection{\@startsection{subsection}{2}{\z@}%
                                     {-3.25ex\@plus -1ex \@minus -.2ex}%
                                     {1.5ex \@plus .2ex}%
                                     {\normalfont\fontfamily{phv}\fontsize{14}{17}\bfseries}}
\renewcommand\subsubsection{\@startsection{subsubsection}{3}{\z@}%
                                    {-3.25ex\@plus -1ex \@minus -.2ex}%
                                     {1.5ex \@plus .2ex}%
                                     {\normalfont\normalsize\fontfamily{phv}\fontsize{14}{17}\selectfont}}
\makeatother

\usepackage{amsmath}
\usepackage{graphicx}
\usepackage{enumerate}
\usepackage{natbib} 
\usepackage{url} 

\usepackage{amsfonts, amsthm, latexsym, amssymb}
\usepackage{lineno}
\usepackage[ruled]{algorithm}
\usepackage{algorithmic}
\usepackage{caption}
\usepackage{subcaption}
\usepackage{cleveref}
\usepackage{multirow}
\usepackage{array}

\newcommand{\mb}{\mathbf}
\newcolumntype{M}[1]{>{\centering\arraybackslash}m{#1}}
\begin{document}
	
\def\spacingset#1{\renewcommand{\baselinestretch}%
	{#1}\small\normalsize} \spacingset{1}

\if0\blind
{
	\title{\bf Partitioned Active Learning for Heterogeneous Systems}
	
	\author{Cheolhei Lee $^a$, Kaiwen Wang $^b$, Jianguo Wu $^c$, Wenjun Cai $^b$, and Xiaowei Yue $^a$ \\
	$^a$ Grado Department of Industrial and Systems Engineering, \\Virginia Tech, Blacksburg, VA, USA \\
    $^b$ Department of Materials Science and Engineering, \\Virginia Tech, Blacksburg, VA, USA \\
    $^c$ Department of Industrial Engineering and Management, \\Peking University, Beijing, China}
	\date{}
	\maketitle
} \fi

\if1\blind
{
    \title{\bf Partitioned Active Learning \\for Heterogeneous Systems}
    \author{Author information is purposely removed for double-blind review}
	
    \bigskip
	\bigskip
	\bigskip
	\begin{center}
		{\LARGE\bf Partitioned Active Learning \\for Heterogeneous Systems}
	\end{center}
	\medskip
} \fi
\bigskip
	
\begin{abstract}
Active learning is a subfield of machine learning that focuses on improving the data collection efficiency of expensive-to-evaluate systems. Especially, active learning integrated surrogate modeling has shown remarkable performance in computationally demanding engineering systems. However, the existence of heterogeneity in underlying systems may adversely affect the performance of active learning. In order to improve the learning efficiency under this regime, we propose the partitioned active learning that seeks the most informative design points for partitioned Gaussian process modeling of heterogeneous systems. The proposed active learning consists of two systematic subsequent steps: the global searching scheme accelerates the exploration of active learning by investigating the most uncertain design space, and the local searching exploits the circumscribed information induced by the local GP. We also propose Cholesky update driven numerical remedies for our active learning to address the computational complexity challenge. The proposed method is applied to numerical simulations and two real-world case studies about (i) the cost-efficient automatic fuselage shape control in aerospace manufacturing; and (ii) the optimal design of tribocorrosion-resistant alloys in materials science. The results show that our approach outperforms benchmark methods with respect to prediction accuracy and computational efficiency.
\end{abstract}
			
	\noindent%

	\spacingset{1.5} 


\section{Introduction} \label{s:intro}
Active learning is a subfield of machine learning and artificial intelligence that maximizes information acquisition to train models data-efficiently \citep{settles.tr09}. Contrary to nonsequential passive learning such as Latin Hypercube design (LHD) and factorial design \citep{santner2018design}, active learning sequentially selects design points in the modeling phase after observing intermediate models and outputs. It is also called query learning, sequential design, adaptive sampling, or optimal design in other literature, while they pursue the same objective: finding the best subset of inputs from the design space according to information criteria that evaluates the informativeness of input referring to uncertainty, disagreement, etc. Active learning has got increasing attention from various applications where sampling is timely and costly demanding, such as quality engineering, response surface investigation, and image recognition \citep{alaeddini2019sequential, cao2020hyperspectral}. Especially, active learning for Gaussian processes (GPs) has been vastly addressed in both methodological perspective and applications, and has shown remarkable performance in practice \citep{seo2000gaussian, gramacy2009adaptive, yue2020active}. 

\begin{figure*}[t!]
\centering
\begin{subfigure}[b]{0.45\textwidth}
\centering
\includegraphics[width=\textwidth]{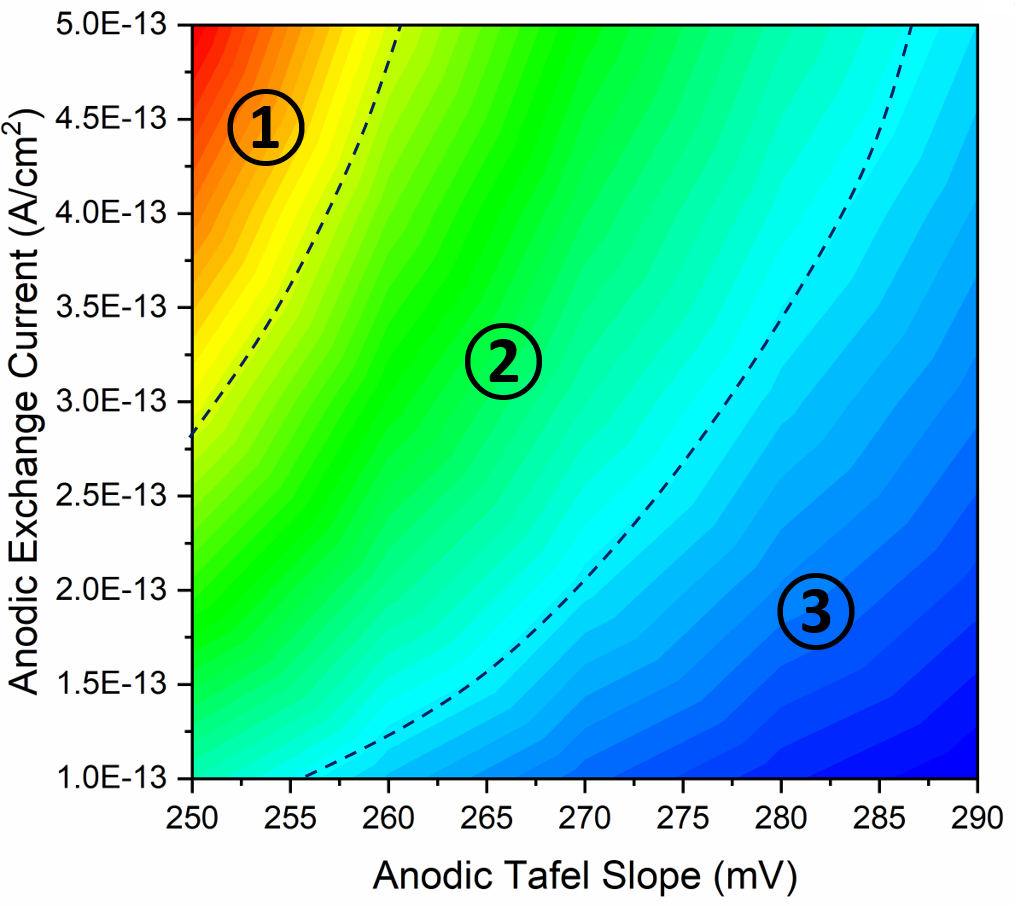}
\end{subfigure}
\hfill
\begin{subfigure}[b]{0.42\textwidth}
\centering
\includegraphics[width=\textwidth]{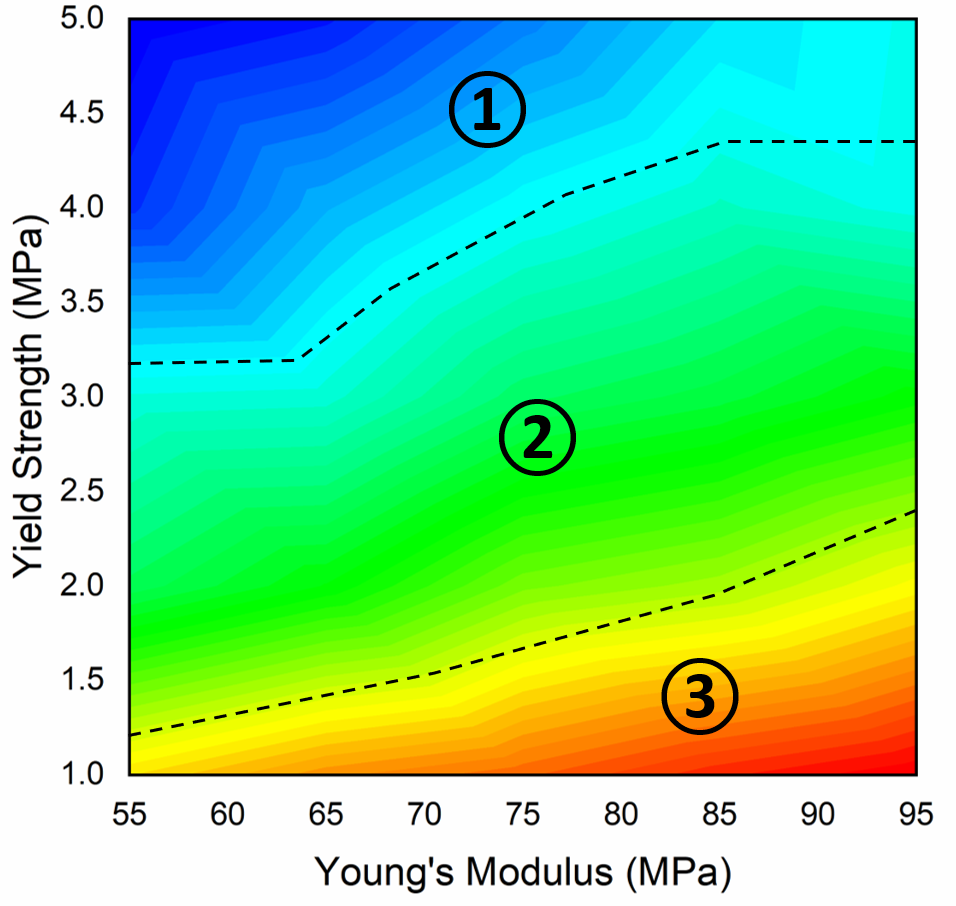}
\end{subfigure}
\caption{Corrosive rates of alloys with different control variables, \textbf{Left:} response surface with Tafel slope and anodic exchange current, \textbf{Right:} response surface with Young's modulus and yield strength.}
\label{fig:tribo}
\end{figure*}

Gaussian processes have been extensively utilized for modeling of various systems spanning robotics, aerospace, and manufacturing process \citep{deisenroth2013gaussian, yue2018surrogate, cho2019hierarchical} due to the capability of uncertainty quantification (UQ) and simplicity \citep{rasmussen2006gpml}. Also, the UQ and diversity of kernel choice in GPs have enabled development of various active learning algorithms, while they have been mostly studied with single GPs (SGPs) that may be inappropriate for systems with heterogeneity (e.g., discontinuity, and abrupt variations in gradient norms or frequencies). Such heterogeneity is ubiquitous in industrial and engineering systems. For example, composite materials, one of the most versatile materials in various contemporary products, are anisotropic and highly nonlinear to external treatments \citep{hyer2009stress}, so they exhibit different behaviors in the design space \citep{lee2020neural}. Another example is the corrosive rates of aluminum alloys. Fig. \ref{fig:tribo} illustrates the corrosive rates of alloys emulated with the finite element method (FEM) over two pairs of control variables. We can observe that the response surface shows spatial heterogeneity, so the design space can be partitioned into three subregions according to the levels of variation. In both cases, the efficiency of active learning driven by SGPs can be significantly deteriorated, since SGPs cannot learn their heterogeneity data-efficiently, thereby misleading the information criteria in active learning.

Partitioned GPs (PGPs) can overcome the modeling limitations of SGPs in heterogeneous systems by allocating multiple independent local GPs on disjoint subregions. Subregions are defined or estimated according to distinguishable characteristics of target systems, so PGPs can accommodate the heterogeneity with local GPs. Moreover, the partitioning improves the scalability of GPs, one of the main drawbacks, by introducing sparsity in their covariance matrices. Due to these advantages, a few approaches for PGP modeling have been proposed \citep{kim2005analyzing, heaton2017nonstationary, pope2021gaussian}. However, most of them have been utilized without active learning, and the existing algorithms for SGPs are suboptimal for PGPs in terms of learning and computational efficiency. More explicitly, the SGP active learning strategy does not take into account the heterogeneity in local GPs, so they are computationally inefficient and cannot conduct macroscopic exploration. Considering the predominance of heterogeneity in industrial and engineering systems, PGPs are urgently in need of a bespoke active learning.

Motivated from the aforementioned limitations of existing active learning strategies, we propose an active learning method for data acquisition and modeling of heterogeneous systems. The proposed method seeks the most uncertain subregion in the design space with the global searching scheme, and the localized integrated mean squared error (IMSE) criterion is subsequently utilized to select the most informative design point within the subregion. The contributions of this paper are summarized as follows. First, we develop an information criterion for capturing the heterogeneity of response surface. Second, we propose novel active learning with a two-step searching scheme that exploits the partitioned structure of models so as to improve the sampling efficiency for the heterogeneous systems. Third, we propose Cholesky update driven numerical remedies to reduce the computational cost of information criterion in our active learning algorithm.

The remainder of this paper is organized as follows. In Section \ref{s:literature}, we review existing active learning methods for GPs, and discuss several PGP modeling approaches with some active learning strategies therein. In Section \ref{s:method}, we elucidate the new partitioned active learning algorithm, and provide applicable techniques for improving the learning efficiency and numerical costs. In Section \ref{s:simulation}, we implement our method on the approximation of two simulation functions in different dimensions, predictive modeling of residual stress in composite structures and corrosive rates of alloys in Section \ref{s:case}. To evaluate our method in both simulation and case studies, we compare the performance to other benchmark methods. Finally, a brief summary of this paper is provided in Section \ref{s:conclusion}.


\section{Literature Review}\label{s:literature}
In this section, we review related literature of existing active learning strategies for SGPs and recent applications in real-world cases. Afterward, we discuss literature of the PGP method and some active learning schemes therein.

\subsection{\emph{Active Learning for Gaussian Processes}}\label{ss:literature_algp}
Due to the capability of UQ in GPs, the predictive uncertainty has been frequently involved in the construction of active learning criteria for GPs. \cite{seo2000gaussian} compared two criteria for GP regression. Both are suggested by \cite{mackay1992information} and \cite{cohn1996active}, and called active learning Mackay (ALM) and active learning Cohn (ALC), respectively. The ALM refers to the variance-based criterion that selects unobserved points with the largest predictive uncertainty quantified with variance or standard deviation. A number of applications exist that employ the ALM due to its simplicity and straightforwardness \citep{uteva2018active}. Meanwhile, the ALC refers to the integrated mean squared prediction error (IMSE) that seeks a point expected to reduce the most the integrated variance of the GP model over the design space, and it has also been widely applied to different GP frameworks. \cite{chen2017sequential} utilized the IMSE criterion for stochastic kriging \citep{ankenman2010stochastic}, so as to balance between exploration over the design space and exploitation with additional replications. \cite{yue2020active} also proposed two active learning algorithms for GP surrogate models of multimodal systems. The variance-based weighted active learning uses the weighted sum of variance-based criteria of modes, and the D-optimal weighted active learning uses the fisher information matrix with the same engineering-driven weighting procedure. The Bayesian optimization (BO) \citep{snoek2012practical} is another interesting trend of sequential sampling that utilizes the predictive uncertainty of GPs for global optimization problems. Although the BO and active learning show subtle nuance in their objectives, they share the common principle, reduction in sampling cost. There are several criteria (also called acquisition functions) for the BO, such as the expected improvement (EI) and the probability of improvement (PI). They refer to the GP surrogate model of blackbox objective function, thereby choosing a point to be queried most likely to be optimal. 

For strategies that do not take the predictive uncertainty into account or less consider, \cite{pasolli2011gaussian} suggested using space-filling designs in the Kernel Hilbert space induced by intermediate GP models. \cite{freytag2013labeling} and \cite{kading2018active} used expected model output change as their criterion for active learning. The strategy chooses the location where is expected to induce the largest change of model outputs over the design space. \cite{erickson2018gradient} and \cite{marmin2018warped} involve the gradient of GP model's response in active learning so as to draw more samples from where the response abruptly changes. Consequently, the gradient-based criterion focuses more on local variation than the entire design space, thereby reducing the prediction error more efficiently. \cite{uteva2018active} and \cite{kim2009construction} referred to the discrepancy of multiple GP models trained on different subsets of data. However, the aforementioned active learning strategies mostly devised for SGPs that are inappropriate for modeling heterogeneous systems. Moreover, it is inefficient to directly apply them to PGPs, since they cannot consider the partitioned structure of PGP models. \cite{liu2018survey} provided a comprehensive active learning strategy for GP surrogate modeling of engineering systems.

\subsection{\emph{Partitioned Gaussian Processes}}\label{ss:literature_pgp}
Although there are techniques other than partitioning design spaces to overcome the limitation of stationary GP models such as input-dependent length scale, warping and convolution kernels \citep{heinonen2016non, marmin2018warped, higdon1998process}, we mainly focus on PGPs that explicitly partition the design space for multiple GP models. \cite{kim2005analyzing} proposed the piecewise GP using Voronoi tessellation with training dataset for partitioning the design space. Advantages of Voronoi tessellation are simplicity, consistency, and distance-based algorithm that coincides with stationary GPs. They estimated the number of regions and centers with the Monte-Carlo approach, and fitted independent GP models for subregions. Subsequently, \cite{pope2021gaussian} generalized the partitioning by merging convex Voronoi cells in order to generate nonconvex subregions, and relaxed centers of cells. \cite{gramacy2008bayesian} proposed the Treed GP (TGP) using decision trees for partitioning. They fitted independent local GPs on each leave corresponds to partitioned subregions. \cite{heaton2017nonstationary} proposed to partition the design space prior to GP modeling by hierarchical clustering referring to finite differences of samples. They insist that the approach allows avoiding the expensive Markov Chain Monte-Carlo (MCMC)  algorithm in the aforementioned approaches. The mixture of GP experts proposed by \cite{rasmussen2002infinite} is another approach using multiple stationary GPs. Although components in their modeling process have own terminologies, the underlying idea is very close to the aforementioned methodologies. 

For active learning in PGPs, \cite{pope2021gaussian} proposed to an active learning algorithm with their PGP modeling that takes a point within regional boundaries and maximizes the space-filling property of design points. However, it focuses more on detecting discontinuity in design space rather than reducing the prediction error over the design space, although such discontinuity is rare in industrial and engineering applications. \cite{gramacy2009adaptive} and \cite{konomi2014bayesian} provided active learning algorithms that are modified versions of \cite{seo2000gaussian} for the TGP. They provided more elaborate techniques considering the posterior structure of partitioned design space. However, their approaches are highly dependent on the tree classifier, while the tree partitioning mostly induces boundaries parallel to axes that may not be realistic in practice \citep{heaton2017nonstationary, pope2021gaussian}. Moreover, the choice of design point candidates are dependent to the areas of partitioned regions, so it can be irrelevant when a plausible partition is not realizable with a few tree-partitioned regions. If we expand our field of interest to classification with generic machine learning models, \cite{cortes2019region} has a similar concept of ours in the macroscopic aspect, while there are main differences: (i) they consider classification problems, thus theoretical bounds they provided may be intractable in practice with regression problems (e.g., model capacity, bounded loss); and (ii) they consider the disagreement-based criterion within a model class, while we advocate the IMSE criterion, which belongs to variance reduction criterion.


\section{Partitioned Active Learning}\label{s:method}
In this section, we propose our active learning algorithm for the PGP modeling. First, we briefly discuss a generic framework of PGPs as the initialization step of active learning, and elucidate our strategy consecutively. Lastly, applicable remedies by Cholesky update are proposed for improving the computational complexity of the algorithm. 

As mathematical notations, we use lower letters for scalars, and distinguish vectors with boldface. Upper letters indicate sets or matrices, and a set of subregion indices are denoted as $[M]=\{1,\ldots, M\}$. In subscriptions, we use parentheses numbers for indices of subregions, and the subsequent normal letter as the indices of letter in the set or matrix. If the index of subregion is apparent, we may omit the region index and only use the index in the set.

\subsection{\emph{Surrogate Modeling for Heterogeneous Systems}}\label{ss:method_init}
We aim to build a PGP model of a heterogeneous system which is expensive to operate. Suppose that the system can be expressed as a target function $f$ defined over a design space $\Omega\subset \mathbb{R}^d$ mapping to $\mathbb{R}$, and the function is endowed with the heterogeneity in $\Omega$. Let $h:\Omega\to \mathbb{R}$ be our estimated PGP model of $f$, and $g:\Omega\to [M]$ be the region classifier that partitions $\Omega$ into $m$ disjoint regions such that $\Omega = \bigcup_{m=1}^M \Omega_{(m)}$ accordance with the heterogeneity. Then, the PGP composed of $M$ local GPs is
\begin{align}
    h(\mb{x}) = \sum_{m=1}^M 1_{\{g(\mb{x})=m\}}h_{(m)}(\mb{x}),\qquad\mb{x} \in \Omega,
\end{align}
where $1_{\{C\}}$ is an indicator function which has value 1 when $C$ is true and 0 otherwise, and $h_{(m)}$ is a local GP assigned to $\Omega_{(m)}$. Although PGPs can take any valid kernel, we mainly consider stationary kernel family such as radial basis function (RBF) and Mat\'ern. Suppose any $\mb{x},\, \mb{x}'\in \Omega$, and $\mb{x}=\left[\:x_1\ \ldots\ x_d\:\right]^\top$. The local GP defined over $\Omega_{(m)}$ with the RBF kernel is defined as
\begin{align}
    h_{(m)}(\mb{x}) \sim \mathcal{GP}\left(\mu_{(m)}(\mb{x}),\ k_{(m)}(\mb{x},\, \mb{x}')\right),\quad \mb{x},\, \mb{x}'\in \Omega_{(m)},
\end{align}
\begin{align}
    k_{(m)}(\mb{x},\, \mb{x}')=\tau_{(m)}^2\prod_{j=1}^{d}\exp\left(-\frac{(x_j- x_j')^2}{l_{(m), j}^2}\right) + \sigma^2_{(m)}1_{\{\mb{x}=\mb{x}'\}},
\end{align}
where $\mu_{(m)}(\mb{x})$ is the mean function assumed to be 0 without loss of generality, and $k_{(m)}$ is the kernel in which nonnegative $\tau_{(m)},\ l_{(m), j}$ and $\sigma^2$ are referred as scale, length and noise hyperparameters. Note that $\mb{l}_{(m)} = \left[\:l_{(m), 1}\ \ldots\ l_{(m), d}\:\right]^\top$ implies that the local GP implements automatic relevance determination (ARD) \citep{rasmussen2006gpml}. We use $\Theta$ as the set of hyperparameters of PGP, that is $\Theta=\{\tau,\, \mb{l},\, \sigma\}$, and $\Theta_{(m)}=\{\tau_{(m)},\, \mb{l}_{(m)},\, \sigma_{(m)}\}$.

Suppose we have finite $n$ observations on $X=\{\mb{x}_1,\ldots, \mb{x}_n\}\subset \Omega$ such that $D = \{(\mb{x}_1, y_1), \ldots, (\mb{x}_n, y_n)\}$, where $y_i$ is the observation at $\mb{x}_i$ with noise, $= f(\mb{x}_i) + \epsilon$ and $\epsilon \sim \mathcal{N}(0,\ \sigma^2)$. According to the region classifier, $X$ and $D$ can be partitioned as $X_{(m)} = \{\mb{x}_{(m), 1}, \ldots, \mb{x}_{(m), n}\}$, and $D_{(m)} = \{(\mb{x}_{(m), 1}, y_{(m), 1}), \ldots, (\mb{x}_{(m), 1}, y_{(m), 1})\}$. Let the covariance matrix associated with $X_{(m)}$ be $K_{(m)}$ such that $(K_{(m)})_{ij} = k_{(m)}(\mb{x}_{(m),i},\, \mb{x}_{(m),j})$ for $i,\,j\in[n]$. The PGP model can be trained with $D$ by maximizing the log marginal likelihoods of local GPs, which is proportional to
\begin{align}
    \ell(\Theta) \propto \sum_{m=1}^M \left(\mb{y}_{(m)}^\top K_{(m)}^{-1} \mb{y}_{(m)} - \log\det{K_{(m)}}-n_{(m)}\log{2\pi}\right)\label{eq:likelihood},
\end{align}
where $\mb{y}_{(m)} = \left[\:y_{(m),1}\ \ldots\  y_{(m),n}\:\right]^\top$, and $n_{(m)}$ is the number of samples in $D_{(m)}$. Note that the PGP's likelihood is the multiplication of that of local GPs due to their independence. Consequently, possibly with some ordering process, the PGP produces a block diagonal covariance matrix which implies that the numerical advantage of PGPs comes from the sparsity. Although the construction of entire covariance matrix is generally unnecessary in practice, it informs us that the model can be manipulated more efficiently by treating each local GP independently.

The performance of PGP is determined by the credibility of the region classifier $g$ and estimated hyperparameters of local GPs $\boldsymbol{\Theta}$. When the underlying partitioning rule is unknown, $g$ must be estimated, which is the major focus of most PGP approaches. The MCMC approach is generally employed, while the expensive MCMC inference can be avoided by clustering finite differences \citep{heaton2017nonstationary} and input-output pairs \citep{nguyen2004active}. We assume that the region classifier $g$ is already available or can be estimated via one of the aforementioned approaches, thereby focusing more on the estimation of $\Theta$. Since $\Theta$ would be inferred by maximizing Equation \eqref{eq:likelihood} given $D_n$, which is generally realized with nonsequential designs, it is unwise to exhaust the sampling budget at the first state. Hence, we take only a small portion in the initial step.

Unfortunately, there is no universal concrete theorem for the optimal portion of initial sampling, while some empirical suggestions can be found in \citep{liu2018survey, yue2020active}. However, we can conjecture that the number of initial samples has a trade-off property. If the initial sampling is weighed too much, the advantage of active learning will be diluted. Otherwise, active learning can be hindered by low-quality information come from unreliable intermediate models. Also, when the region classifier $g$ is not explicitly known so that $g$ must be estimated with the initial samples, the number should be enough to obtain an acceptable $g$.

\subsection{\emph{Partitioned Active Learning Strategy}}\label{ss:method_pal}
The essence of active learning is an information criterion function $J:\Omega\to \mathbb{R}$ that quantifies the potential importance of a design point. By optimizing $J(x)$ with respect to $x\in\Omega$, the next design point is determined and queried to the target function, which is also called the oracle in active learning. The variance-based and the IMSE criteria are mostly considered in active learning for GPs due to their versatility and simplicity, so we shortly summarize two criteria and tailor them to establish a new criterion for PGPs.

Suppose we have $D \subset \Omega$, and intend to determine the next location $\mb{x}_{n+1}\in\Omega$ with GPs without partitioning. The variance-based criterion is defined as
\begin{align}
    J_{\text{VAR}}(\mb{x}) = \text{Var}_n(\mb{x}) := k(\mb{x}) - \mb{k}(\mb{x},\, X)^\top K^{-1} \mb{k}(\mb{x},\, X)\label{eq:var_based},
\end{align}
where $\text{Var}_n(\mb{x})$ is the predictive variance at $\mb{x}$ given $n$ observations $D$. The ALM maximizes the variance-based criterion so as to select the location with the greatest predictive uncertainty. Meanwhile, the IMSE criterion is calculated by
\begin{gather}
    J_{\text{IMSE}}(\mb{x}) = \int_{\Omega}\text{Var}_{n+1}(\mb{s}|\mb{x})\,p(\mb{s})\,d\mb{s},\label{eq:IMSE}\\
    \text{Var}_{n+1}(\mb{s}|\mb{x}) = k(\mb{s}) - \mb{k}(\mb{s},\, X_{n+1})^\top K_{n+1}^{-1}\,\mb{k}(\mb{s},\, X_{n+1}), \quad X_{n+1} = \left[\:X_n\ \  \mb{x}\:\right]^\top, \nonumber
\end{gather}
where $\mb{s}\in\Omega$ with density (or importance) $p(\mb{s})$, and $K_{n+1}$ is the covariance matrix associated with $X_{n+1}$. Minimizing the IMSE criterion selects the location which is expected to reduce the predictive uncertainty the most over $\Omega$, and we refer the active learning with the IMSE criterion as the ALC.

There are more behind derivations of both criteria, while we mention shortly herein. It turns out that the ALM is equivalent to the maximum entropy design, since the choice leads to maximizing the determinant of covariance matrix given $\Theta_n$. Meanwhile, the ALC can be explained by minimizing the generalization error of statistical learning, which can be decomposed into the famous form,
\begin{align}
    \mathbb{E}_\Omega\left[\|f-h\|^2\right] = \mathbb{E}_\Omega\left[\|f-h^\ast\|^2\right] + \mathbb{E}_\Omega\left[\|h^\ast-h\|^2\right] \nonumber,
\end{align}
where $h^\ast$ is the best estimation of $f$ in the given PGP class. Assuming that the first term representing the bias of the best model is acceptable, the ALC focuses on minimizing the second term, which is the variance. Regarding the variance of $h^\ast$ as a constant, it becomes minimizing the overall variance of $h$.

Although the variance-based criterion is more straightforward and numerically inexpensive than the IMSE criterion, the ALC empirically has shown better performance than the ALM \citep{seo2000gaussian, gramacy2009adaptive}. Moreover, one of the promising characteristic of the ALC is that it avoids sampling from the boundary of design space which may provoke loss of information, while the ALM frequently does. It makes the IMSE criterion more appealing to PGPs, since local GPs share common boundaries, so the variance-based criterion may lead to oversampling near the boundary shared with two adjacent local GPs. Another advantage of the ALC is that it is possible to consider the importance of $\mb{s}\in\Omega$ in the information criterion, while the ALM cannot. It allows us to incorporate prior knowledge and to give more weight on a specific region, thereby making the algorithm more distinguishable from the space-filling design.

However, the IMSE criterion of \eqref{eq:IMSE} does not consider multiple local GPs simultaneously, thus the block diagonal structure of covariance matrix is not involved. Therefore, we dedicate to construction of our criterion for the PGPs by aggregating every information criterion attained by each local GPs. When the IMSE criterion is considered for a candidate location $\mb{x}\in \Omega_{(m)}$ with PGPs, it can be written as
\begin{align}
    J(\mb{x}|\mb{x}\in \Omega_{(m)}) =& \sum_{i\ne m}\int_{\Omega_{(i)}}\text{Var}_{(i)}(\mb{s}_{(i)})\,p(\mb{s}_{(i)})\,d\mb{s}_{(i)}\nonumber\\& + \int_{\Omega_{(m)}}\text{Var}_{(m), n_{(m)}+1}(\mb{s}_{(m)}|\mb{x})\,p(\mb{s}_{(m)})\,d\mb{s}_{(m)},\label{eq:IMSE_pal}
\end{align}
where $\mb{s}_{(m)}\in \Omega_{(m)}$ with density $p(\mb{s}_{(m)})$. The interpretation of each term in \eqref{eq:IMSE_pal} is quite worthwhile. The first term is the sum of IMSEs except for $h_{(m)}$, and it is invariant to the choice of $\mb{x}\in \Omega_{(m)}$. The second term is equivalent to \eqref{eq:IMSE}, in which $h_{(m)}$ is only considered, so that \eqref{eq:IMSE_pal} will only take account of the local region associated with the candidate location. Consequently, there are two main differences between \eqref{eq:IMSE} and \eqref{eq:IMSE_pal}: (i) consideration of IMSEs over other local regions; and (ii) the localized IMSE criterion. We focus on each term subsequently considering their meanings, thereby efficiently minimizing \eqref{eq:IMSE_pal}. 

Heuristically, \eqref{eq:IMSE_pal} is more likely to be minimized when the most uncertain local GP is taken account into the second term, since the local GP has more potential to be reduced with additional observations. Each IMSE in the first term indicates the regional uncertainty of PGP, so it can be used for investigating the most uncertain region. We involve every local GP in the first term of \eqref{eq:IMSE_pal}, and conduct the global searching prior to calculating the second term by using the following criterion
\begin{align}
    J_\text{G}(m) &= \int_{\Omega_{(m)}}\text{Var}_{(m), n_{(m)}}(\mb{s}_{(m)})\,p(\mb{s}_{(m)})\,d\mb{s}_{(m)}, \nonumber\\
    m^\ast &= \arg\max_{m\in[M]} J_G(m).\label{eq:global_IMSE}
\end{align}
A possible problem can be aroused in the global searching is that if multiple regions have similar uncertainties, then this scheme will only care about the most uncertain one. However, it can be easily resolved by taking a percentile or providing a threshold to consider multiple regions in the global searching. 

Once the most uncertain region is determined by \eqref{eq:global_IMSE}, we focus on the second term within $\Omega_{(m^\ast)}$ as the local searching, which is
\begin{align}
    J_\text{L}(\mb{x}) &= \int_{\Omega_{(m^\ast)}}\text{Var}_{(m^\ast), n_{(m^\ast)}+1}(\mb{s}_{(m^\ast)}|\mb{x})\,p(\mb{s}_{(m^\ast)})\,d\mb{s}_{(m^\ast)},\nonumber\\
    \mb{x}^\ast &= \arg\min_{\mb{x}\in \Omega_{(m^\ast)}} J_L(\mb{x}).\label{eq:local_IMSE}
\end{align}
Since the local GP in \eqref{eq:local_IMSE} reflects the local behavior of underlying function excluding heterogeneity from other regions, it can lead to improvement in exploitation of active learning by avoiding implausible predictive uncertainty. Moreover, it turns out that the localized IMSE criterion reduces the computational cost of IMSE criterion due to the reduced size of local GP. We call the sequential criteria \eqref{eq:global_IMSE} and \eqref{eq:local_IMSE} by Partitioned IMSE (PIMSE), and the active learning with PIMSE as Partitioned ALC (PALC). 

Two termination criteria can be considered in the algorithm. One is based on the pre-defined budget, and the other is the achievement of desired prediction accuracy with the model. The budget criterion is straightforward, while most active learning algorithms may not guarantee that the last model is the best one unless the trained dataset is noise-free. To avoid the unwanted overfitting, early-stopping with a desired prediction accuracy is one possible solution. Note that, to utilize the prediction accuracy criterion for the termination, a separated testing dataset or cross validation scheme is required for the model evaluation in the learning process.

\begin{algorithm}[!t]
\caption{PALC}
\begin{algorithmic}[1]\label{algo:palc}
\STATE \textbf{Prerequisite}: $M, N_0, N_{\text{Max}}, N_{\text{Ref}}, N_{\text{Cand}}, \Omega, (\text{Optional: }L, D_{\text{Test}}, e^\ast)$
\STATE Initialize $D_0$ with $N_0$ space-filling design over $\Omega$
\STATE Partition $\Omega$ into $M$ subregions
\STATE Train PGP ($h$) on $D_0$
\WHILE{card($D$) $< N_{\text{Max}}$ and $L_{D_{\text{Test}}}(h)> \epsilon^\ast$}
\STATE Generate $X_{\text{Ref}}\subset \Omega$ with $N_{\text{Ref}}$ space-filling design
\STATE \textbf{Global Searching: }Solve $m^\ast = \arg\max_{m} J_\text{G}(m)$
\STATE Generate $X_{\text{Cand}}\subset\Omega_{(m^\ast)}$ with $N_{\text{Cand}}$ space-filling design
\STATE \textbf{Local Searching: }Solve $\mb{x}^\ast = \arg\min_{\mb{x}\in X_{\text{Cand}}} J_\text{L}(\mb{x})$
\STATE Observe $y^\ast$ at $\mb{x}^\ast$
\STATE $D = D\cup \{(\mb{x}^\ast, y^\ast)\}$
\STATE Train $h$ on $D$
\IF{$D_{\text{Test}}$ exists}
\STATE Check $L_{D_{\text{Cand}}}(h) < e^\ast$
\ENDIF
\ENDWHILE
\end{algorithmic}
\end{algorithm}

The pseudocode of PALC is provided in Algorithm \ref{algo:palc}. In prerequisites, $(L, D_{\text{Test}}, e^\ast)$ is optionally required for the use of predictive accuracy criterion which of $L$ is a loss function, $D_{\text{Test}}$ is a separated dataset, and $e^\ast$ is a desired prediction accuracy. $N_0$ and $N_{\text{Max}}$ stand for the numbers of initial samples and the maximum attainable samples (i.e., budget), respectively. $X_{\text{Ref}}$ is a reference set composed of $N_{\text{Ref}}$ space-filling points ($\{\mb{s}_i\}_{i=1}^{N_{\text{Ref}}}\subset\Omega$), which is required to implement the global searching approximately as
\begin{align}
    J_\text{G}(m) = \frac{1}{N_{Ref}}\sum_{i=1}^{N_{Ref}} \text{Var}_{(m), n_{(m)}}(\mb{s}_i)\,\tilde{p}(\mb{s}_{(m), i}),\nonumber
\end{align}
where $\tilde{p}(\mb{s}_{(m), i})$ is the approximated probability mass function at $\mb{s}_{(m), i}$. The subset of $X_{\text{Ref}}$ composed of points in $\Omega_{(m^\ast)}$, which is $\{\mb{s}_i\}_{i=1}^{N_{\text{Ref}}^\ast}$ will be referred to the approximated local searching subsequently as
\begin{align}
    J_\text{L}(\mb{x}) = \frac{1}{N_{\text{Ref}}^\ast}\sum_{i=1}^{N_{\text{Ref}}^\ast}\text{Var}_{(m^\ast), n_{(m^\ast)}+1}(\mb{s}_{(m^\ast), i}|\mb{x})\,\tilde{p}(\mb{s}_{(m^\ast), i}),\label{eq:approx_local}
\end{align}
where $\mb{x}\in X_{\text{Cand}}=\{\mb{x}_i\}_{i=1}^{N_{\text{Cand}}}$. In this paper, we have generated new $X_{\text{Ref}}$ and $X_{\text{Cand}}$ with LHD in every step in order to encourage exploration.

\subsection{\emph{Cholesky Update based Numerical Remedies to Tackle the Computational Complexity Challenge}}\label{ss:method_numerical}
The IMSE criterion is numerically more demanding than the variance-based criterion, since it involves the inversion of $K_{n+1}$ which should be updated for every candidate. That is, calculation of the IMSE criterion requires $\mathcal{O}(n^3)$ for each candidate. Moreover, when $N$ candidates are provided to the active learning module, the computational cost is multiplied by the number. Although the significance of their effects may vary with situations, the effect of candidate number can be more considerable than the inversion cost. Therefore, in order to improve the numerical aspect of PALC, we should provide some remedies for both matrix inversion and the number of candidates.

The global searching can be used for reducing the number of candidates by narrowing down the region of interest. Generally, candidates for active learning are given with space-filling or dense-grid over the design space. Therefore, given that candidates are provided uniformly over the overall design space, taking the subset of candidates in the most uncertain region with the global searching leads to reduction in the number of matrix inversions proportional to
\begin{align*}
    \text{Pr}(\mb{s}\in\Omega_{(m^\ast)}) = \int_{\Omega_{(m^\ast)}}p(\mb{s}_{(m^\ast)})\,d\mb{s}_{(m^\ast)},
\end{align*}
which is the ratio of the most uncertain region considering the importance of input. 

The matrix inversion cost of PGP is automatically alleviated by partitioning the design space with the block diagonal covariance matrix. That is, the inversion cost reduces from $\mathcal{O}(n^3)$ to at most $\mathcal{O}({n_{(m)}}^3)$, where ${n_{(m)}} < n$ usually. Another applicable remedy is updating the inverse of $K_{n_{(m)}+1}$ in \eqref{eq:local_IMSE} exploiting $K_{n_{(m)}}^{-1}$ iteratively. Although it is possible to apply the Sherman-Morrison formula to get the updated inverse matrix directly as shown in \cite{gramacy2009adaptive}, we propose to use the Cholesky decomposition \citep{rasmussen2006gpml} for solving the linear system $K_{n_{(m)}}^{-1}\mb{k}$ considering the numerical stability and cost. Given that the Cholesky factor (a lower triangular matrix) of $K_{n_{(m)}}$ is known, the update of Cholesky factor of $K_{n_{(m)}+1}$ with that of $K_{n_{(m)}}$ only requires the forward substitution step of the size $n_{(m)}$ triangular system, thus it needs only $\mathcal{O}(n_{(m)}^2)$ instead of $\mathcal{O}({n_{(m)}}^3)$. A more detailed procedure of the Cholesky update is provided as follows.

\subsubsection*{\emph{Updating Cholesky Factor for PIMSE Criterion}} \label{appendix:cholupdate}
Suppose we have the Cholesky factor $L_n$ of $K_n$, which is the covariance matrix of $X_n$, such that $K_n = L_n {L_n}^\top$. We aim to get the Cholesky factor $L_{n+1}$ of
\begin{align}
    K_{n+1} = \left[\begin{array}{c c} K_n & \mb{k_n^\ast} \\ \mb{k_n^\ast}^\top & k(\mb{x}^\ast)\end{array}\right]\nonumber,
\end{align}
where $\mb{x}^\ast$ is a candidate input, and $\mb{k}_n^\ast = k(X_n,\, \mb{x}^\ast)$. Since $K_{n+1}$ shares the same part of $K_n$, it turns out that $L_{n+1}[:n,:n]$ is equivalent to $L_n$. For $L_{n+1}$, we can apply the Cholesky-Banachiewicz algorithm for $i=1, \ldots, n$ as
\begin{gather}
    L_{n+1, i} = L_{i, i}^{-1}\left(\mb{k_n^\ast}_{i} - \sum_{j=1}^{i-1}L_{n+1, j}L_{i, j}\right),\nonumber\\
    L_{n+1, n+1} = \sqrt{k(\mb{x}^\ast) - \sum_{j=1}^n L_{n+1, j}}.\nonumber
\end{gather}

Rather than calculating the PIMSE directly, the PALC can be faster by skipping redundant computation applying the Cholesky updating approach. The Cholesky factor $L_n$ can be used for the predictive variance of GP in \eqref{eq:var_based} for $\mb{s}\in X_\text{Ref}$, and the global searching criterion of \eqref{eq:global_IMSE} as
\begin{gather}
    \text{Var}_n^2(\mb{s}) = k(\mb{s}) - {\mb{v}_n}^\top {\mb{v}_n},\qquad
    \mb{v}_n = L_n^{-1} {\mb{k}_n},\nonumber
\end{gather}
where $\mb{k}_n = k(X_n,\, \mb{s})$. In the same manner, the optimal solution of the PIMSE criterion \eqref{eq:local_IMSE} can be obtained by minimizing
\begin{gather}
    \text{Var}_{(m^\ast), n_{(m^\ast)}+1}^2(\mb{s}) = k(\mb{s}) - \mb{v^\ast}^\top \mb{v}^\ast,\label{eq:fast_IMSE1}\\
    \mb{v}^\ast = L_{n+1}^{-1} k(X_{n+1},\, \mb{s}),\label{eq:fast_IMSE2}
\end{gather}
Since we already have the solution of \eqref{eq:fast_IMSE2} partially with $\mb{v}$ (i.e., $\mb{v}^\ast[1:n] \equiv \mb{v}_n$), we need only $v^\ast_{n+1}:=\mb{v}^\ast[n+1]$, which can be calculated with a forward substitution as
\begin{align}
    v^\ast_{n+1} = k(\mb{x}^\ast, \mb{s}) - \sum_{j=1}^n L_{n+1, k}\mb{v}_{j}\nonumber.
\end{align}
If the importance $p(\mb{s})$ over $\Omega$ is trivial (e.g., uniform), we can simplify \eqref{eq:fast_IMSE1}. Since $k(\mb{s})$ in \eqref{eq:fast_IMSE1} is invariant to $\mb{s}$ and $\mb{x}^\ast$, we can only consider the second term $\mb{v^\ast}^\top \mb{v}^\ast$ as the simplified PIMSE criterion, which must be maximized in the PALC.


\section{Simulation Study}\label{s:simulation}
In this section, we evaluate our active learning algorithm with simulation data. Two functions are considered that can be visualized straightforwardly. Both functions contain heterogeneous response surfaces, and observation noise is involved. In order to reduce the variability from the different initialization settings, each simulation is replicated ten times. As our benchmark methods, the ALC and the ALM are considered for SGPs, and the following partitioned active learning methods are considered: (i) the variance-based criterion for the local searching referred to as PALM; (ii) PALC without global searching (PALC-NoG). Additionally, uniform random sampling (Rand) and LHD are also considered as passive learning. For evaluation of models, one thousand space-filling design points for each design space are used with root mean squared error (RMSE) as a metric. Computational times spent on considered methods are also compared excluding the labelling and model training times.

\subsection{\emph{One-dimensional Data}}\label{ss:simulation_1d}
\begin{figure*}[t!]
\centering
\begin{subfigure}[b]{0.48\textwidth}
\centering
\includegraphics[width=\textwidth]{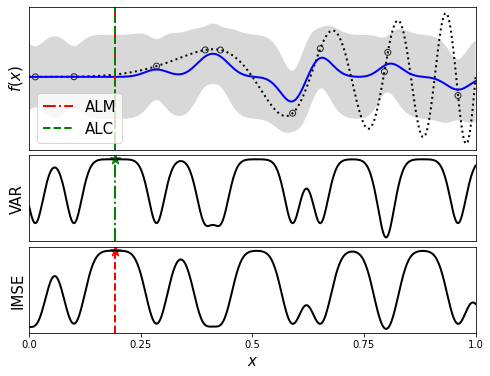}
\end{subfigure}
\hfill
\begin{subfigure}[b]{0.48\textwidth}
\centering
\includegraphics[width=\textwidth]{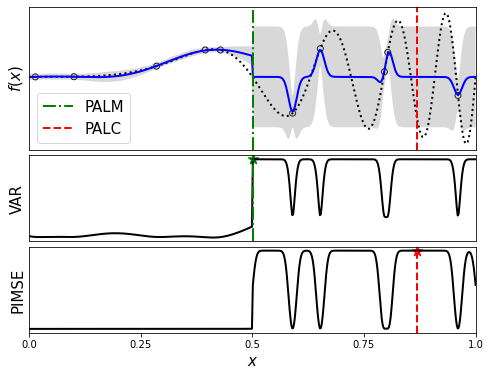}
\end{subfigure}
\caption{Results of different active learning algorithms in 1-D simulation, \textbf{Left:} SGP, \textbf{Right:} PGP with two local GPs}
\label{fig:1d_sim}
\end{figure*}

We apply our proposed active learning algorithm to a one-dimensional simulation function
\begin{align}
    f(x) = 2x\sin{(8\pi x^3)}\nonumber,
\end{align}
which is defined on [0, 1] as the dotted line in Fig. \ref{fig:1d_sim}. The observation noise is imposed with zero-mean Gaussian noise whose variance is $\sigma^2=10^{-4}$. We allocate ten samples for initial training using Maximin LHD, and 20 samples are sequentially obtained via active learning. The function is differentiable, with heterogeneous frequency and amplitude over the domain. Although there is no specific boundary for this function, we rigorously partition the design space with the following translated heaviside function, $g(x)=\mb{1}_{\{x\ge 0.5\}}$, which is assumed as the ground truth.

Fig. \ref{fig:1d_sim} shows each GP model fitted with initial samples. First, the right figure in Fig. \ref{fig:1d_sim} illustrates that the PGP prevents misled active learning by providing appropriate predictive uncertainty. Also, we can observe from the left of Fig \ref{fig:1d_sim} that the next design point to be queried in the SGP model is chosen from the low frequency region with both variance-based and IMSE criteria. Meanwhile, in the PGP model, the variance-based criterion takes the point from the boundary, while the IMSE does not. It implies that the IMSE criterion can be more promising when the adjacent local GPs show comparable predictive uncertainties.

After full running of each active learning strategy, the learning curves of considered models are presented in the left plot of Fig. \ref{fig:sim_curve}. From the result, we can observe that both partitioned active learning strategies outperform the SGPs, and the PALC is slightly better than the PALM. Considering the first query location of each method in Fig. \ref{fig:1d_sim}, it is not so surprising that the PALC performs the best. Table \ref{tab:sim1} summarizes the results after the full data acquisition, and we grouped considered methods into three: (i) passive learning (Rand; and LHD); (ii) variance-based methods (ALM; and PALM); and (iii) IMSE-based methods (ALC; PALC-NoG; and PALC). We can observe that the predictive accuracy of PALC outperforms the others. In the computational time, variance-based methods (ALM and PALM) are definitely faster than the IMSE-based methods, while they are deficient in predictive accuracy; the ALM even worse than the random sampling and the LHD. Among the IMSE-based methods, the PALC is faster than the others. Moreover, if we focus on the PALC and PALC-NoG, we can observe that the global searching does not only reduce the computational time, but also improves the learning efficiency.

\begin{table}[ht]
\caption{One-dimensional simulation study results}
\centering
\begin{tabular}{M{0.3\textwidth}|M{0.3\textwidth}M{0.3\textwidth}}
\hline
Methods       & RMSE                   & Time (s)                \\ \hline
Rand          & 0.395 (0.152) & -                      \\
LHD           & 0.332 (0.133) & -                      \\ \hline
ALM           & 0.352 (0.159) & 0.341 (0.140)           \\
PALM          & 0.072 (0.270) & 0.144 (0.049)           \\ \hline
ALC           & 0.272 (0.205) & 11.468 (0.424)          \\
PALC-NoG      & 0.051 (0.277) & 7.049 (0.024)          \\
\textbf{PALC} & \textbf{0.048 (0.273)} & \textbf{3.553 (0.137)} \\ \hline
\end{tabular}
\label{tab:sim1}
\end{table}

\subsection{\emph{Two-dimensional Data}}\label{ss:simulation_2d}
\begin{figure*}[t!]
\centering
\includegraphics[width=\textwidth]{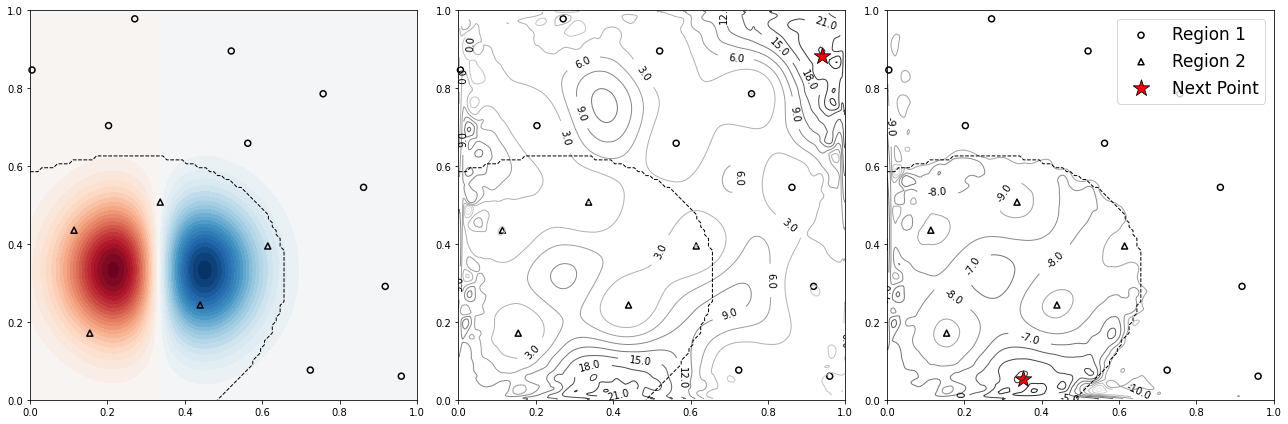}
\caption{IMSE criterion contour plots in 2-D  simulation, \textbf{Left:} ground truth and initial design points with partitioned regions, \textbf{Center:} IMSE of ALC, \textbf{Right:} PIMSE of PALC}
\label{fig:2d_sim}
\end{figure*}

We expand our active learning experiment on a two-dimensional function (shown in the left of Fig. \ref{fig:2d_sim}), which is also used in \citep{gramacy2009adaptive, konomi2014bayesian}. Since the purpose of the function is to evaluate the performance of the proposed active learning approach, it is comprised of two regions: even and uneven, and zero-mean Gaussian noise with variance $\sigma^2=10^{-6}$ is imposed as the observation noise. In a similar manner, we begin with 15 samples with LHD, and obtain 15 additional samples via active learning.

For partitioning the design space, we borrow some ideas from \cite{heaton2017nonstationary}, exploiting the finite differences between initial samples. Unlikely to their original clustering approach, we use a support vector classifier (SVC) after labeling initial samples based on their magnitudes of finite differences to generate a more flexible boundary than the Voronoi tessellation. Consequently, we obtain a partitioning shown in the left figure of Fig. \ref{fig:2d_sim}. Since the region 1 is less interesting than the region 2, we can observe that the PIMSE criterion induced by two independent local GPs provides more relevant information of design points as shown in the right plot of Fig. \ref{fig:2d_sim}. Consequently, the IMSE criterion with the SGP fails to pick from the more interesting region due to the misled information criterion.

\begin{figure*}[t!]
\centering
\begin{subfigure}[b]{0.48\textwidth}
\centering
\includegraphics[width=\textwidth]{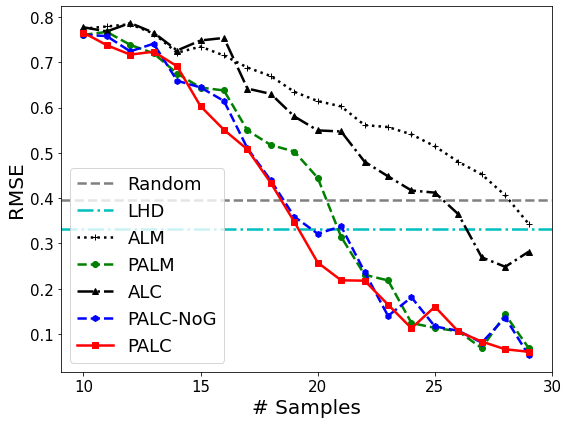}
\end{subfigure}
\hfill
\begin{subfigure}[b]{0.48\textwidth}
\centering
\includegraphics[width=\textwidth]{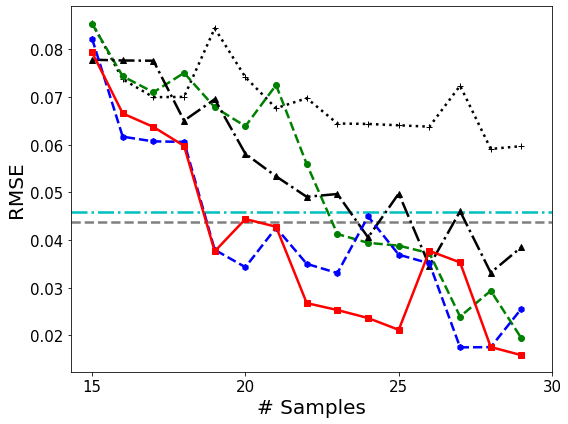}
\end{subfigure}
\caption{Learning curves in simulation study, \textbf{Left:} one-dimensional, \textbf{Right:} two-dimensional}
\label{fig:sim_curve}
\end{figure*}

In Table \ref{tab:sim2}, we can see that the PALC surpasses the other methods again in both predictive accuracy and computational time among the IMSE-based methods. Also, the learning curve of PALC in the right plot of Fig. \ref{fig:sim_curve} shows that the PALC is more promising than the active learning with SGPs.

\begin{table}[ht]
\caption{Two-dimensional simulation study results}
\centering
\begin{tabular}{M{0.3\textwidth}|M{0.3\textwidth}M{0.3\textwidth}}
\hline
Methods       & RMSE                   & Time (s)                \\ \hline
Rand          & 0.044 (0.017) & -                       \\
LHD           & 0.046 (0.021) & -                      \\ \hline
ALM           & 0.060 (0.023) & 0.344 (0.081)           \\
PALM          & 0.019 (0.025) & 0.151 (0.026)           \\ \hline
ALC           & 0.039 (0.013) & 8.368 (0.161)          \\
PALC-NoG      & 0.026 (0.004) & 5.570 (0.593)          \\
\textbf{PALC} & \textbf{0.016 (0.005)} & \textbf{1.924 (0.016)} \\ \hline
\end{tabular}
\label{tab:sim2}
\end{table}

\section{Case Study}\label{s:case}
In this section, we apply our approach to construction of surrogate models of two real-world cases. The original purpose of surrogate models herein is to embed them into the automatic shape control system of fuselage assembly process, and the optimal design of aluminum alloy.  Both cases include higher input dimensions than the previous simulations, and heterogeneity in their design spaces due to complex material properties.

\subsection{\emph{Residual Stress of Composite Fuselages in Aerospace Manufacturing}}\label{ss:case_composite}
We apply our proposed active learning strategy to the construction of predictive model of residual stress in the composite fuselage assembly process. In the aircraft manufacturing process, composite fuselages are built in several subsections independently, so they are subject to the discrepancy in the junction part. In consequence, the reshaping of composite fuselage is conducted with multiple fixed actuators with the automatic shape control. In order to achieve the optimal manufacturing process, the shape control needs to consider not only the deformation, but also the residual stress of the structure due to their fatal affects on the final product. The development of highly accurate predictive model for the shape control is very challenging, since the problem is endowed with both the heterogeneity and the demanding cost of real experiments and physics-based model. Especially, the stress of composite fuselage is more difficult than deformation to predict \citep{lee2020neural}, we apply our method and other benchmarks to predictive modeling of the stress.

\begin{figure*}[t!]
\centering
\begin{subfigure}[b]{0.48\textwidth}
\centering
\includegraphics[width=0.92\textwidth]{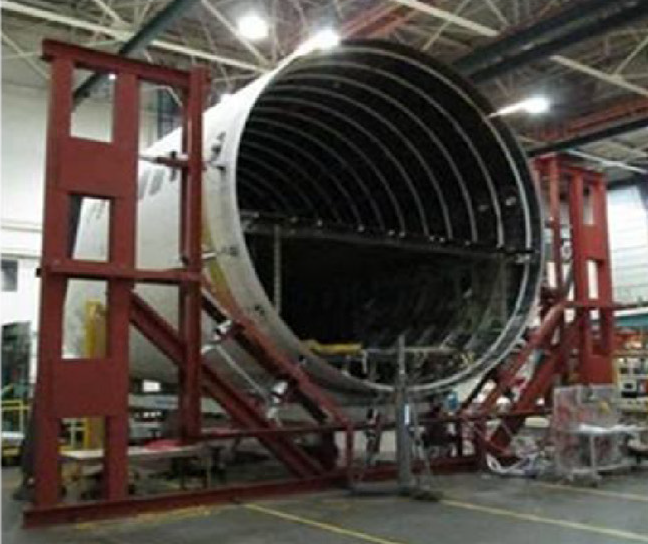}
\end{subfigure}
\hfill
\begin{subfigure}[b]{0.48\textwidth}
\centering
\includegraphics[width=\textwidth]{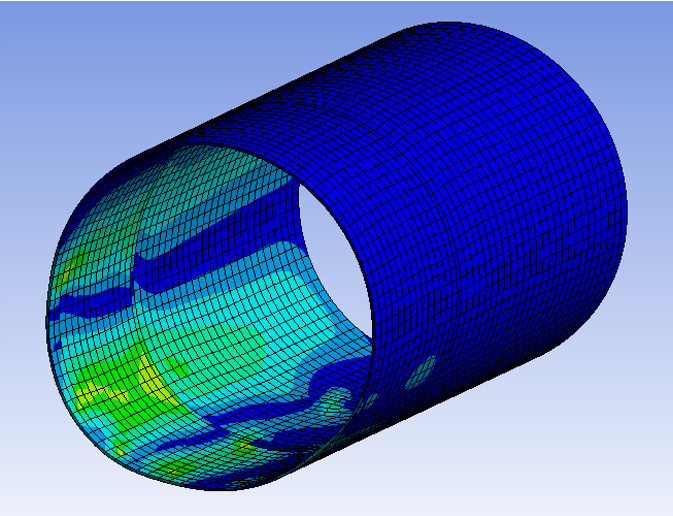}
\end{subfigure}
\caption{Shape adjustment of composite fuselage, \textbf{Left:} composite fuselage installed upon the fixture with actuators, \textbf{Right:} simulated residual stress of composite fuselage in ANSYS}
\label{fig:case1}
\end{figure*}

In order to implement our case study cost-efficiently, we utilized the FEM model, which is well-calibrated based on the real experiment \citep{wang2020effective}. The simulation mimics the real shape adjustment process that has ten actuators under the fuselage section as shown in the left figure of Fig. \ref{fig:case1} \citep{wen2018feasibility}, and the maximum magnitude of actuator's force is 450 lbf. The maximum residual stress on the fuselage section is our interest, which is measured in the psi scale. The ten-dimensional design space is partitioned into three regions using SVC based on the clustering of input and output pairs, and the ARD Mat\'ern kernel is used for each local GPs and SGPs. As the initial dataset, 20 samples are given with the LHD, and additional 20 samples are sequentially obtained from different one thousand of different LHD points with different active learning strategies. The model evaluation is conducted with a separated testing dataset composed of 100 LHD samples, and mean absolute error (MAE) is used as a metric.

Table \ref{tab:case1} summarizes the results of each learning strategy in the case study. The PALC surpasses the other methods in both predictive error and the computational time among the IMSE criterion-based methods. Interestingly, except for the PALC-NoG, the other active learning methods are worse than two passive learning strategies. Moreover, the PALM performs the worst in this case, even though they chose design points to spread out the design space well. In computational times, we can observe that the PALC-NoG has taken more time than the ALC. This is due to the domination of one subregion in this case. More explicitly, the PALC and PALC-NoG have assigned 15 and 16 design points of 20 additional samples from one region, thereby decreasing the efficiency of our matrix inversion scheme. However, it also shows that the global searching of PALC scheme has reduced not only the computational time, but also the predictive error.

\begin{table}[ht]
\caption{Residual stress case study results}
\centering
\begin{tabular}{M{0.3\textwidth}|M{0.3\textwidth}M{0.3\textwidth}}
\hline
Methods       & MAE                   & Time (s)                \\ \hline
Rand          & 3.858           & -                       \\
LHD           & 3.319          & -                       \\ \hline
ALM           & 4.545          & $2.6\times10^{-4}$           \\
PALM          & 11.555          & $2.5\times10^{-4}$           \\ \hline
ALC           & 4.693          & 11.596          \\
PALC-NoG      & 3.691          & 12.750          \\
\textbf{PALC} & \textbf{3.207} & \textbf{9.729} \\ \hline
\end{tabular}
\label{tab:case1}
\end{table}

\subsection{\emph{Tribocorrosion in Aluminium Alloys}}\label{ss:case_tribo}
\begin{figure}[t!]
    \centering
    \includegraphics[width=0.4\textwidth]{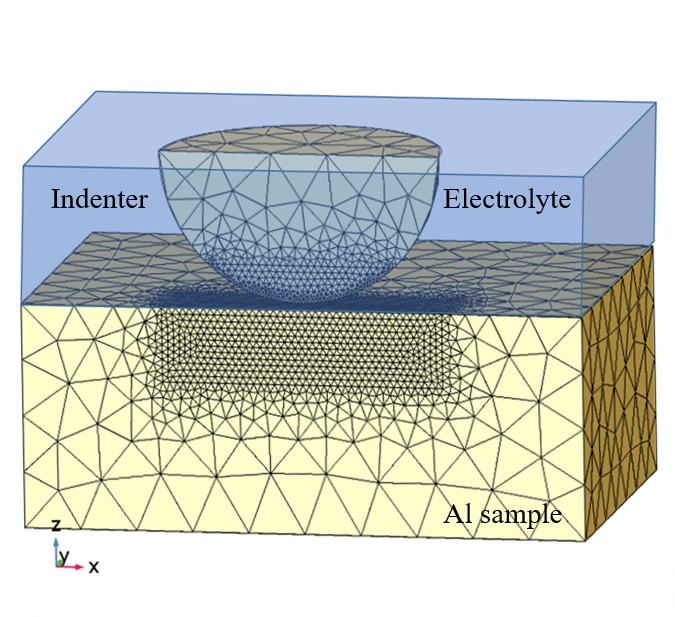}
    \caption{Schematic tribocorrosion simulation setup.}
    \label{fig:tribo_set}
\end{figure}

As our second case study, the material loss rate during stress corrosion (i.e. tribocorrosion) in aluminum alloys with six control variables are considered. To test the tribocorrosion resistance of metals, experimental tests and FEM simulations were carried out by scratching the surface of the samples in corrosive environment \citep{wang2021multiphysics, wang2021modeling} as shown in Fig. \ref{fig:tribo_set}. During the tribocorrosion process, the mechanical deformation and the electrochemical processes including active corrosion and passivation work synergistically to cause material degradation. The FEM model calculates the contact mechanics between the indenter and the sample and simulate the wear process as well as the wear-accelerated material dissolution of the corrosion process, and generate the volume loss results. The 6 control variables for the FEM model are material property descriptors: young's modulus, yield strength, anodic Tafel slope, anodic exchange current density, cathodic Tafel slope, and cathodic exchange current density. The former two govern the mechanical properties while the latter four determine the corrosion behavior of the alloy. The output of the FEM model is the tribocorrosion rate of the alloy, expressed as volume loss per time.

The surrogate model of the FEM model is constructed to assist the optimal design of alloys with uncertainty quantification and alleviating the high-computational cost of the FEM model. To establish the relationship between material property and tribocorrosion rate, a total of 106 FEM simulations were performed by systematically varying the 6 control variables. Since scales of variables in the dataset are inconsistent, each variable is normalized to be within a unit interval. For evaluation, mean relative error (MRE) is used as a metric due to the infinitesimal scale of the output. The MRE is calculated as 
\begin{align}
    \text{MRE} = \frac{1}{n}\sum_{i=1}^n \frac{|y_i - h(x_i)|}{|y_i|}\nonumber.
\end{align}

The PGP in this case is composed of 3 local GPs with the ARD RBF kernel, and the SVC model is used for partitioning based on the normalized input-output pairs. Considering the relatively small size of samples, 5-fold cross-validation is used. That is, about 84 samples are passed to each model as the candidate set, and the rest of samples are used for the model evaluation. Compared methods are trained up to 50 samples from 20 common initial samples.

Table \ref{tab:case2} shows the results of each learning strategy in which standard deviations are obtained from the cross-validated evaluation. We can observe that the PALC achieves the minimum averaged predictive error and the computational time among the IMSE-based methods. Passive learning methods are worse than others, and the variance-based methods are also worse than the IMSE-based methods. In this case, overall computational times are much lower than the previous case due to the small number of samples in the candidate pool, while the numerical remedies in PALC have significantly reduced the time of ALC.

\begin{table}[ht]
\caption{Tribocorrosion case study results}
\centering
\begin{tabular}{M{0.3\textwidth}|M{0.3\textwidth}M{0.3\textwidth}}
\hline
Methods       & MRE                   & Time (s)                \\ \hline
Rand          & 0.028 (0.017)          & -                       \\
LHD           & 0.028 (0.017)          & -                       \\ \hline
ALM           & 0.026 (0.017) & 0.014 (0.003)           \\
PALM          & 0.023 (0.012) & 0.104 (0.000)           \\ \hline
ALC           & 0.022 (0.013) & 1.283 (0.127)          \\
PALC-NoG      & 0.022 (0.012) & 0.764 (0.018)          \\
\textbf{PALC} & \textbf{0.020 (0.013)} & \textbf{0.757 (0.017)} \\ \hline
\end{tabular}
\label{tab:case2}
\end{table}


\section{Conclusion}\label{s:conclusion}
Active learning is a special machine learning that seeks to improve sampling efficiency and lower data collection cost. Existing active learning strategies mainly focus on investigating one response surface, so they are insufficient for reliable and cost-efficient surrogate modeling of heterogeneous systems. This paper dedicated establishing an efficient partitioned active leaning strategy that adopts two-step searching schemes based on the partitioned IMSE criterion structure. By partitioning the design space into multiple subregions according to heterogeneity in the target system, the global searching scheme refers to the integrated predictive uncertainties of local GPs to determine the most uncertain subregion. The global searching scheme allows us to reduce the region of interest, thereby not only accelerating the searching speed, but also improving the overall learning efficiency as shown in the simulation and case studies. The local searching scheme exploited the chosen local GPs in the global searching phase, so the localized IMSE criterion may provide more relevant information minimizing the interruption of heterogeneous characteristics in the other regions.

For the numerical perspective of active learning, the following applicable remedies are provided: reducing the number of candidates with the global searching scheme, and the Cholesky factor update, which can be embedded into the PALC. From two of each simulation and case study, the proposed method outperformed the benchmark methods including passive learning, the variance-based methods, and the IMSE-based methods. Moreover, the global searching scheme dramatically improved the performance of PALC by comparing our method to the PIMSE without the global searching step. The results of simulation and case studies imply that it is also applicable to other domains where heterogeneity exists.

\if0\blind{
\section*{Acknowledgements}
The authors acknowledge the financial support by the US National Science Foundation CMMI-2035038 and CMMI-1855651.
} \fi

\bibliographystyle{chicago}
\spacingset{1}
\bibliography{arxiv_pal}
	
\end{document}